\documentclass[journal]{IEEEtran}

\usepackage{todonotes}
\usepackage{biblatex}
\usepackage{booktabs}
\usepackage{amsmath}
\usepackage{caption}
\usepackage{subcaption}
\usepackage{multirow}
\AtBeginBibliography{\footnotesize}

\begin{document}

\title{Towards Fingerprint Mosaicking Artifact Detection: \\A Self-Supervised Deep Learning Approach}

\author{Laurenz~Ruzicka,
        Alexander~Spenke,
        Stephan~Bergmann,
        Gerd~Nolden,
        Bernhard~Kohn;
        Clemens~Heitzinger
\thanks{L. Ruzicka and B. Kohn are with the Department
of Digital Safety and Security, Austrian Institute of Technology, Vienna.\\
E-mail: firstname.lastname@ait.ac.at.}
\thanks{A. Spenke, S. Bergmann, and G. Nolden are with the Federal Office for Information Security, Bundesamt für Sicherheit in der Informationstechnik, Bonn.\\
E-mail: firstname.lastname@bsi.bund.de
}
\thanks{C. Heitzinger is with the Center for Artificial Intelligence and Machine Learning (CAIML) and Department of Computer Science, TU Wien, Vienna.\\
E-mail: Clemens.Heitzinger@TUWien.ac.at}%
}


\maketitle

\begin{abstract}
Fingerprint mosaicking, which is the process of combining multiple fingerprint images into a single master fingerprint, is an essential process in modern biometric systems. However, it is prone to errors that can significantly degrade fingerprint image quality. This paper proposes a novel deep learning-based approach to detect and score mosaicking artifacts in fingerprint images. Our method leverages a self-supervised learning framework to train a model on large-scale unlabeled fingerprint data, eliminating the need for manual artifact annotation. The proposed model effectively identifies mosaicking errors, achieving high accuracy on various fingerprint modalities, including contactless, rolled, and pressed fingerprints and furthermore proves to be robust to different data sources. Additionally, we introduce a novel mosaicking artifact score to quantify the severity of errors, enabling automated evaluation of fingerprint images.
By addressing the challenges of mosaicking artifact detection, our work contributes to improving the accuracy and reliability of fingerprint-based biometric systems.
\end{abstract}

\IEEEpeerreviewmaketitle

\section{Introduction}

\IEEEPARstart{F}{ingerprints} have long been established as a critical biometric trait for personal identification and verification, owing to their uniqueness and permanence \cite{monson_permanence_2019}. They play an essential role in a wide range of applications, from law enforcement and border control to unlocking personal devices and securing sensitive information. The accuracy and reliability of fingerprint-based systems hinge on the quality of the collected fingerprint images.

Fingerprints are typically collected through one of two primary methods: contact-based \cite{ali_overview_2016} or contactless \cite{conti_improving_2023} approaches. In contact-based methods, a user's finger is physically pressed or rolled on a scanner, capturing detailed ridge and valley structures. Contactless methods, on the other hand, capture the fingerprint image without direct contact, using cameras or other sensors. In particular, in the process of creating rolled fingerprints for both modalities, multiple partial images must be stitched together in a process known as mosaicking to create a comprehensive representation of the rolled fingerprint \cite{cui_dense_2021, fengling_mosaic_2014, chen_successive_2019}. The challenge of creating mosaicking artifact free representations is known from several disciplines like photography, medicine or even AI 3D scene generation \cite{adel_image_2014, engstler_invisible_2024, samsudin_development_2013}.  

However, the mosaicking process is prone to errors, which can significantly compromise the integrity of the resulting fingerprint image by shifting minutiae position or even creating new minutiae \cite{liu_touchless_2013}. As a consequence, these errors lead in worst case scenarios to misidentification or non-identification. A well-known example of misidentification, based on a latent fingerprint, is the case of Brandon Mayfield \cite{cole_more_2005}.

In this work, we propose to classify these errors into two categories: soft and hard mosaicking errors. Soft mosaicking errors refer to deformations within the fingerprint that maintain continuous junctions of ridge lines to the neighboring regions. These mostly require a reference fingerprint for detection, because such deformations could also be inherent to the fingerprint.  Hard mosaicking errors, on the other hand, are characterized by visibly incorrect alignments in the ridge-valley structures. In some cases, a discontinuous junction is hidden by blurring or alpha blending the discontinuity of the misaligned patch, which can lead to in-between cases where the classification into a soft or hard mosaicking error is more difficult. Measures such as local sharpness \cite{vu_s-3_2012} can reveal these ambiguous cases.


Given the critical impact of relative minutiae positions on fingerprint recognition accuracy, there is a pressing need for a reliable method to identify hard mosaicking errors. Currently, to the best of our knowledge, no framework or tool exists that can detect hard mosaicking errors in fingerprint images.
Early detection of mosaicking artifacts can enable prompt re-acquisition or flagging of erroneous data, ensuring that only high-quality fingerprint images are used for subsequent analysis and comparison. This paper introduces a novel fingerprint mosaicking artifact detector based on a self-supervised deep learning approach, designed to address this need and enhance the robustness of fingerprint-based biometric systems.

\subsection{Related Work}

\textbf{Fingerprint Mosaicking}

Fingerprint mosaicking aims to combine multiple fingerprint impressions to create a complete and accurate master fingerprint image. Various methods have been developed to address the challenges associated with this process.

Jain and Ross (2002) presented an early approach to fingerprint mosaicking, emphasizing the importance of accurate registration and alignment to improve recognition performance \cite{jain_fingerprint_2002}. They provided fundamental insights into the complexities of mosaicking multiple fingerprint images.
Choi et al. (2005) explored fingerprint mosaicking through rolling and sliding techniques, which helps in capturing larger fingerprint areas by combining multiple smaller impressions \cite{choi_fingerprint_2005}. 
Ross, Shah, and Shah (2006) compared image-based and feature-based mosaicking methods, concluding that while both approaches have merits, the choice depends on the specific application requirements \cite{ross_image_2006}. Their comparative analysis provided a broader perspective on different mosaicking strategies.
Liu et al. (2013) focused on contactless fingerprint (CL) acquisition and mosaicking, addressing the challenges of capturing high-quality fingerprints without direct contact \cite{liu_touchless_2013}. This approach is particularly relevant for enabling the contactless sensor modality.
Bhati and Pati (2015) introduced a fingerprint mosaicking algorithm using the phase correlation method, which focuses on aligning fingerprint patches accurately to form a seamless composite image \cite{bhati_novel_2015}.
Chen et al. (2019) proposed a successive minutia-free mosaicking technique for small-sized fingerprints, which does not rely on minutiae points. This technique marked a shift from traditional minutiae-based methods to more comprehensive alignment strategies \cite{chen_successive_2019}.
Cui, Feng, and Zhou (2021) developed a dense registration and mosaicking method using an end-to-end deep learning network, which significantly improves the accuracy of the mosaicked fingerprint by learning from a large dataset of fingerprint images \cite{cui_dense_2021}. Their method represents the cutting edge of leveraging deep learning for fingerprint mosaicking.

\textbf{Reference Free Error Detection:}
On the other side, a lot of work was put into detecting mosaicking errors in various domains. For example, Nabil et al. \cite{nabil_error_2017} focused on detecting stitching errors in panoramic videos. They proposed a pairwise assessment technique to identify inconsistencies and misalignment during the stitching process, which is analogous to detecting hard mosaicking errors in fingerprint images.
Zhang et al. \cite{zhang_subjective_2017} discussed both subjective and objective methods for evaluating the quality of panoramic videos. Their approach includes assessing visual artifacts and distortions that can also be related to detecting hard mosaicking errors in fingerprint mosaicking.
Conze et al. \cite{conze_objective_2012} addressed the evaluation of synthesized views' quality. Their focus on objective metrics to assess view synthesis quality parallels the need to objectively detect and evaluate hard mosaicking errors in fingerprint mosaicking.
Vu et al. \cite{vu_s-3_2012} introduced a spectral and spatial measure of local perceived sharpness in images. This measure can be adapted to detect masked hard mosaicking errors in fingerprints by identifying areas with inconsistent sharpness, indicating possible deformations.

\textbf{Reference Based Error Detection:}
Battisti et al. \cite{battisti_objective_2015} discussed objective image quality assessments of 3D synthesized views, focusing on objective metrics to evaluate the geometric and visual accuracy of synthesized images. This methodology can be applied to fingerprint mosaicking, when reference images are used to detect and quantify mosaicking errors.
The National Institute of Standards and Technology (NIST) defined "Geometric Accuracy" \cite{libert_guidance_2018} as a critical parameter in evaluating the fidelity of synthesized views. Applying this concept to fingerprint mosaicking involves comparing the geometric accuracy of the mosaicked fingerprint against a reference to identify errors.
Libert et al. \cite{libert_1d_2009} proposed a 1D spectral image validation verification metric for fingerprints, which assesses the spectral properties of fingerprint images to ensure their accuracy and consistency. This metric can detect discrepancies in mosaicked fingerprints by comparing them to reference images.
Zhou et al. discussed structural similarity based image quality assessment \cite{wang_structural_2005}, a method for assessing image quality based on structural similarity to reference images. This approach is highly relevant to detecting both soft and hard mosaicking errors in fingerprints by comparing the structural integrity of the mosaicked image to that of a reference.

\subsection{Contribution}

In this publication, we add to the existing work by proposing a novel framework for detecting hard fingerprint mosaicking artifacts. We provide the following in this work:
\begin{itemize}
    \item Proposal of a novel deep learning model for hard stitching artifact detection, including a self-supervised data annotation pipeline with extensive data augmentation.
    \item Evaluation of model performance on out-of-distribution sensor modalities.
    \item Analysis of model robustness when trained on different modalities and dataset sizes and the analysis of robustness when evaluated on synthetic fingerprints with quality alterations simulating wounds, scars and more.
    \item Proposal of a novel mosaicking artifact score.
    \item Assessment of the impact of mosaicking errors on identification and authentication performance.
\end{itemize}
\section{Methods}

\subsection{Data}

The data used for training our deep learning model was acquired from a real police setting using a contactless sensor, as detailed in Weissenfeld et al. \cite{weissenfeld_case_2022}. The dataset includes 245,193 images from 539 users for training, 30,650 images for validation, and 30,649 images for testing. The training data consists of single-shot, contactless fingerprint images without mosaicking artifacts, serving as the ground truth.

For calculating the impact of the removal of artifacts, we determined the Equal Error Rates (EER) using a subset of the dataset, which includes 500 images from 50 different fingers. 

For testing with different modalities, we utilized the NIST Special Publication 300a \cite{fiumara_nist_2018}, which consists of a set of rolled fingerprint recordings and a corresponding set of slap recordings. Furthermore, it acts as an open benchmark dataset to compare future methods with ours.

In addition, we trained the same architecture twice, with different training data, for robustness and performance evaluation. The data used for the second run was collected from pressed fingerprint  (PR) recordings. They were recorded using a frustrated total internal reflection (FTIR) based fingerprint scanner in a laboratory environment with controlled capture conditions. In total 32,800 fingerprints from 288 identities were captured. From this total, 80\% (26240) images were used for training, 10\% (3280) were used for validating the model during training and 10\% were used as a test set to score the model. We call this total dataset PRD-1 and the test set PRD-1-Test. We assume that no mosaicking artifacts are introduced while capturing pressed fingerprints.

Additionally, rolled fingerprints from the same subjects are captured as well, utilizing the same fingerprint scanner. This dataset is called ROD-1. 

Furthermore, data collection continued after the training run of the model was started, resulting in a new FTIR based pressed fingerprint dataset, called PRD-2. It consists of an additional 5,495 fingerprints of 143 subjects. From those 143 subjects, 117 are also present in the PRD-1 dataset. 

Furthermore, we used a subset of the fingerprint modification approaches of \cite{priesnitz_syncolfinger_2022} applied to 100 images generated with SFinGe \cite{maltoni_synthetic_2009} for the robustness analysis. We generated data simulating different levels of fingerprint alterations, such as skin problems, ink stains and contrast variations, sensor noise, scar creation and wounds.

\subsection{Model}

We developed a self-supervised deep learning model to address the task of detecting fingerprint mosaicking artifacts. In self-supervised learning, the model generates its own supervisory signal from the data, which allows us to leverage large amounts of unlabeled fingerprint images assumed to be free of mosaicking artifacts, such as those captured in a single-shot manner, for training. This approach enables the model to learn robust representations directly from the data without requiring extensive labeled datasets.

\subsubsection{Data Pre-Processing}

In the data pre-processing step, all fingerprint images were resized to a fixed size of $224 \times 224$ pixels. This size was selected to balance detail preservation and computational efficiency, and because it matches the dimensions used in the pre-training of the backbone model.

\subsubsection{Data Augmentation}

To enhance the robustness and generalization capability of our model, the following data augmentation techniques were applied to the fingerprint images before introducing the mosaicking artifacts:

\begin{itemize}
    \item Random Resizing and Cropping
    \item Random Horizontal Flips
    \item Random Rotations
    \item Random Perspective Changes
    \item Gaussian Blurring
    \item Random Solarization
    \item Random Posterization
    \item Random Histogram Equalization
\end{itemize}

These augmentation techniques were applied in random order. 

\subsubsection{Self-Supervised Learning}

To introduce the supervisory signal, we employed two methods of adding mosaicking artifacts. The first method simulates patch based artifacts and the second one line based artifacts along the whole image width or height. The first involves selecting a patch within the image, sized between 5\% and 15\% of the image dimension, and offsetting its pixels by 2\% to 7\% of the image dimensions. We chose relative size constraints and pixel offsets, because they stay independent of the image resolution and can therefore be applied to images with varying dots-per-inches.

\begin{figure}[htb]
    \centering
    \begin{subfigure}[b]{0.33\linewidth}
        \includegraphics[width=0.95\textwidth]{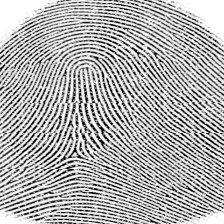}
    \end{subfigure}%
    \begin{subfigure}[b]{0.33\linewidth}
        \includegraphics[width=0.95\textwidth]{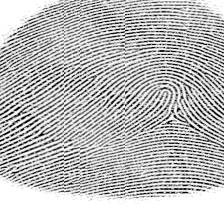}
    \end{subfigure}%
    \begin{subfigure}[b]{0.33\linewidth}
        \includegraphics[width=0.95\textwidth]{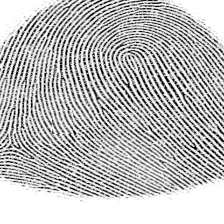}
    \end{subfigure}
    \begin{subfigure}[t]{0.33\linewidth}
        \includegraphics[width=0.95\textwidth]{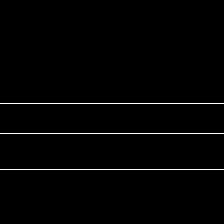}
        \caption{Horizontal Line}
        \label{fig:train_data_artifact_hz}
    \end{subfigure}%
    \begin{subfigure}[t]{0.33\linewidth}
        \includegraphics[width=0.95\textwidth]{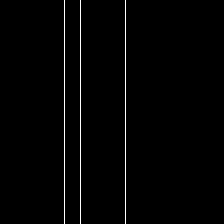}
        \caption{Vertical Line}
        \label{fig:train_data_artifact_vl}
    \end{subfigure}%
    \begin{subfigure}[t]{0.33\linewidth}
        \includegraphics[width=0.95\textwidth]{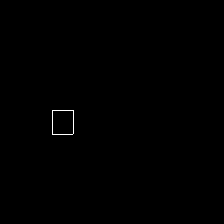}
        \caption{Patch}
        \label{fig:train_data_artifact_patch}
    \end{subfigure}%
    \caption{Depiction of different artifact  types created for supervisory signal}
    \label{fig:train_data_artifact_creation}
\end{figure}

This process, additionally ensuring non-overlapping patches, is repeated up to four times. The second method, chosen with a 25\% probability, involves shifting pixels along a randomly chosen vertical or horizontal line, again up to four times, without mixing horizontal and vertical artifacts within the same image. Figure \ref{fig:train_data_artifact_creation} depicts examples of created artifacts and the corresponding training signal on synthetic data for visualization purposes. Figure \ref{fig:train_data_artifact_hz} shows an example of three horizontal line artifacts, where the fingerprint is shifted along the y-axis. Figure \ref{fig:train_data_artifact_vl} shows three vertical line artifacts and \ref{fig:train_data_artifact_patch} one patch like artifact.

These artificially created artifacts eliminate the need for manual labeling and enable the use of large, unlabeled datasets for training.

\subsubsection{Architecture and Hyperparameter}

\begin{figure}
    \centering
    \includegraphics[width=0.8\linewidth]{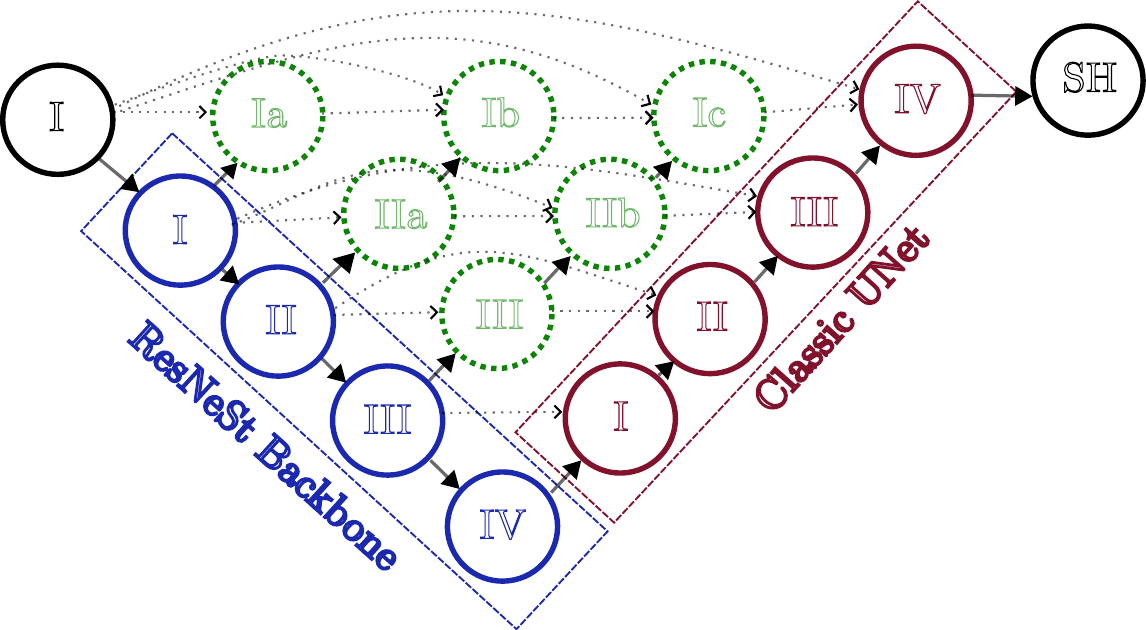}
    \caption{Model architecture of our proposed combination of ResNeSt for the encoder and UNet++ for the general model architecture and decoder design.}
    \label{fig:model_architecture}
\end{figure}

Our model architecture uses the torch segmentation models framework\cite{iakubovskii_segmentation_2019} and leverages the UNet++ design, integrating a ResNeSt-50d encoder pretrained on ImageNet. Figure \ref{fig:model_architecture} provides an abstract visualization of the architecture, highlighting the encoder backbone in blue on the left and the UNet++ decoder in green and red on the right. The circles represent different layers, with their colors grouping them as part of the encoder, decoder, or auxiliary layers. Dashed lines illustrate skip connections, while downward-pointing arrows indicate spatial downsampling and upward-pointing arrows signify spatial upsampling.

The model's input is initially processed through convolution and pooling layers, represented by the black circle with an 'I' at the top left. It then undergoes feature extraction within the ResNeSt encoder. Intermediate outputs from the ResNeSt layers are fed into the UNet++ decoder through two mechanisms:
\begin{enumerate}
    \item they are connected to decoder layers with matching spatial resolution, similar to the skip connections in the original UNet architecture by Ronneberger \cite{ronneberger_u-net_2015}, and
    \item they are upsampled and passed into intermediate processing layers (shown as green dashed layers), which are also linked to the decoder stack via skip connections.
\end{enumerate}%
After traversing the entire encoder and decoder stacks, including all intermediate layers, the processed input reaches the final segmentation head, indicated by the black circle with 'SH' on the right. The segmentation head then produces the predicted segmentation mask as model output.

\paragraph{Encoder}

The ResNeSt-50d (Resilient Split-Attention Networks) architecture takes the role of the encoder in the UNet++ architecture. ResNeSt introduces a split-attention mechanism that enhances feature representation by combining the benefits of ResNet with selective kernel networks. This allows the model to focus on the most relevant features for the task, improving performance on image recognition and segmentation tasks.

\paragraph{UNet++}

UNet++ is an advanced version of the original U-Net architecture, designed for more precise segmentation. It features a nested and dense skip pathway that bridges the encoder and decoder more effectively. This structure enhances the model's ability to capture fine-grained details and contextual information by improving feature propagation and spatial resolution at multiple levels. The architecture includes a series of convolutional layers, upsampling operations, and dense connections that facilitate the learning of intricate patterns in fingerprint images.

\paragraph{Learning Rate and Optimizer}

We set the learning rate to $1.7 \times 10^{-3}$ and employ the Stochastic Gradient Descent (SGD) optimizer with momentum. Momentum is set to 0.9 to enhance the convergence speed and stability of the training process. For the second model trained on PR recordings, we increased the learning rate to $1 \times 10^{-2}$ to speed up the training process.

\paragraph{Loss Function}

We use the Jaccard loss, also known as the Intersection over Union (IoU) loss, which is well-suited for segmentation tasks \cite{nowozin_optimal_2014}. This loss function measures the overlap between the predicted and true segmentation masks, promoting better segmentation performance by directly optimizing the IoU metric.

\paragraph{Training Hyper-Parameters}

The model is trained with a batch size of 64 over 243 epochs. Although we observed no overfitting of the model, as can be seen in the loss plot in the supplementary materials, and still saw a decrease in validation loss, we stopped the training run after 243 epochs. We stopped the run because the validation loss curve showed that no significant gains were to be expected if we had continued the training. The training run took around one month on a Nvidia GeForce RTX 3090. Note however that the biggest bottleneck in this training run was the IO speed at which data could be read. Furthermore, the model can also be trained on a cheaper costumer grade GPU with less VRAM, however this requires a reduced batch size and gradient accumulation. 

For the PR recordings model, we used a Nvidia RTX A6000 GPU to train the model over 201 epochs, which took around twelve hours. 

We implement a custom warm-up strategy for the initial 10 epochs, during which we train on the dataset without image augmentations. This phase utilizes patch sizes that are double the minimum and maximum values, as well as pixel offsets that are also twice the minimum and maximum values. This helps the model to learn the pattern, before increase the difficulty to improve robustness.

\paragraph{Model Size}

The model's architecture consists of three main components: the encoder, the decoder, and the segmentation head. Each component contributes to the overall parameter count and computational complexity, measured in Floating Point Operations (FLOPs). The following Table \ref{tab:model_parameter_flops} provides a detailed breakdown of the parameters and FLOPs for each part of the model:

\begin{table}[htb]
    \centering
    \begin{tabular}{c|cc}
        \toprule
        Part    & Parameter & FLOPs \\
        \midrule
        Encoder & 25.4M & 21.7G\\
        Decoder & 25.5M & 159.8G \\
        Segmentation Head  & 145   & 28.9M \\
        \hline
        Total   & 50.9M & 181.6G \\
        \bottomrule
    \end{tabular}
    \caption{Model Component Breakdown: Parameter Count and Floating-Point Operations (FLOPs). Values are expressed in millions (M) and billions (G).}
    \label{tab:model_parameter_flops}
\end{table}

As seen in Table \ref{tab:model_parameter_flops}, the number of parameters is balanced between encoder and decoder, but the decoder requires more than seven times the compute for its operations. The effect of the final segmentation head on the memory consumption and inference speed can be neglected. 

\subsection{Mosaicking Artifact Score}

In addition to the segmentation mask output, which depicts the regions where mosaicking artifacts were detected, we also introduce a new mosaicking artifact score. This allows for the metric to be used automatically over large datasets, without the need to manually inspect each result. The metric $S$ is calculated as follows:

\begin{align}
    &S := \Biggl( \sum_i^n b_{patch} + w_{patch} \times h_{patch} + \notag \\ 
    &c \times \biggl( \sum_i^m s_{height} \times w_{line} + \sum_i^o s_{width} \times h_{line} \biggr) \Biggr) \times \notag \\
    &\qquad \frac{100}{s_{width} \times s_{height}}, \label{equ:mosaick_artifact_score} \\
    &b_{patch} := b \times  \frac{s_{width} \times s_{height}}{100} \notag
\end{align} 

where $n$ is the number of detected patches where misalignment occurred, $b$ a constant that weights the importance of multiple patches and is set to 5 in this setting, $w_{patch}, h_{patch}$ stand for the width and the height of the detected patch, $c$ is a constant that balances the importance of the single line artifact and is set to 0.025 in our case, $m$ is the number of detected purely vertical line artifacts, $s_{height}$ stands for the height of the segmentation mask, $w_{line}$ is the width of the detected line artifact, $h_{line}$ is the height of the detected line artifact, $o$ depicts the number of detected horizontal line artifacts and $s_{width}$ stands for the width of the segmentation mask.
\section{Experiments and Results}

\subsection{Model Performance}

To evaluate the performance of our CL and PR model in detecting fingerprint mosaicking artifacts, we calculate several key metrics on the test sets, as well as on the slap and rolled data. We use the following metrics:
\begin{itemize}
    \item Intersection over Union (IoU): The IoU score measures the overlap between the predicted segmentation and the ground truth. It is calculated as the ratio of the true positive (TP) pixels to the sum of true positive, false positive (FP), and false negative (FN) pixels. 
    \item Recall: Recall, or sensitivity, measures the proportion of actual positives (artifacts) that are correctly identified by the model. 
    \item F1 Score: The F1 score is the harmonic mean of precision ($TP / (TP+FP)$) and recall, providing a balance between these two metrics. 
    \item F2 Score: The F2 score is similar to the F1 score but gives more weight to recall than precision. 
    \item Accuracy: Accuracy is the ratio of correctly predicted pixels (both true positives and true negatives) to the total number of pixels. 
    \item Mosaicking Artifact Score: Additionally, we also use the newly defined artifact score in equation \ref{equ:mosaick_artifact_score} to calculate the mosaic artifact score for both the label, as well as the model prediction and compare the absolute value of the difference. This is then averaged over the whole dataset.
\end{itemize}


\begin{table*}[t]
    \centering
    \begin{tabular}{cccccccc}
        \toprule
         & Dataset & IoU & F1 & F2 & Accuracy & Recall & Mean Score Dif. \\
         \midrule
         \multirow{3}{0cm}{\rotatebox{90}{CL}} & Weissenfeld et al. \cite{weissenfeld_case_2022} & 0.982 & 0.991 & 0.990 & 1.000 & 0.989 & 0.264\\ 
         & NIST 300a slap & 0.959 & 0.979 & 0.975 & 1.000 & 0.972 & 0.483 \\
         & NIST 300a rolled & 0.908 & 0.952 & 0.940 & 1.000 & 0.932 & 1.061 \\
         \midrule
         \multirow{3}{0cm}{\rotatebox{90}{PR}} & PRD-1-Test & 0.977 & 0.988 & 0.987 & 1.000 & 0.986 & 0.355\\
         & PRD-2 & 0.978 & 0.989 & 0.988 & 1.000 & 0.987 & 0.351\\
         & ROD-1 & 0.931 & 0.964 & 0.957 & 1.000 & 0.952 & 0.815 \\
         \bottomrule
    \end{tabular}
    \caption{Model performance of the with contact-less fingerprints trained model (CL) and the with pressed  fingerprints trained model (PR) measured via intersection-over-union (IoU), F1 and F2 score, accuracy, recall and mean mosaicking artifact score (Mean Score Dif.) for contactless data (first row), contact based rolled data (NIST 300a rolled), contact based slap data (NIST 300a slap), pressed data from the test set (PRD-1-Test), pressed data acquired after the training cutoff (PRD-2) and rolled data (ROD-1).}
    \label{tab:model_performance}
\end{table*}

Table \ref{tab:model_performance} shows the test results on the contactless dataset (Weissenfeld et al. \cite{weissenfeld_case_2022}), the contact based rolled recordings (NIST 300a rolled) and the contact based slap recordings (NIST 300a slap). Furthermore, also the second model trained for the robustness analysis is evaluated on the PRD-1 test set, the PRD-2 dataset and the ROD-1 dataset. The models perform best on the modality that they were trained on, which was contactless fingerprint recordings for the CL model and pressed recordings for the PR model. 

Table \ref{tab:model_performance} furthermore shows the mean mosaicking artifact score difference. The lowest difference in predicted vs. ground truth score can be seen for the data of the modality on which the model was trained on. Similar to the other metrics, the slap data, which is more similar to the contactless data, performed better than the rolled data for the CL model. Also related, the pressed recordings of the PRD-2 dataset performed very similar to the PRD-1 and the largest change in the mean artifact mosaicking score difference can be seen when comparing the results of the pressed data to the ones of the rolled fingerprint recordings. Nonetheless, even the rolled modality stays below a mean score difference of one, indicating good model performance and therefore significantly below the detection threshold of the patch weight $b$ of 5.

\begin{figure}[htb]
    \centering
    \begin{subfigure}{\linewidth}
    \centering
    \includegraphics[width=\linewidth]{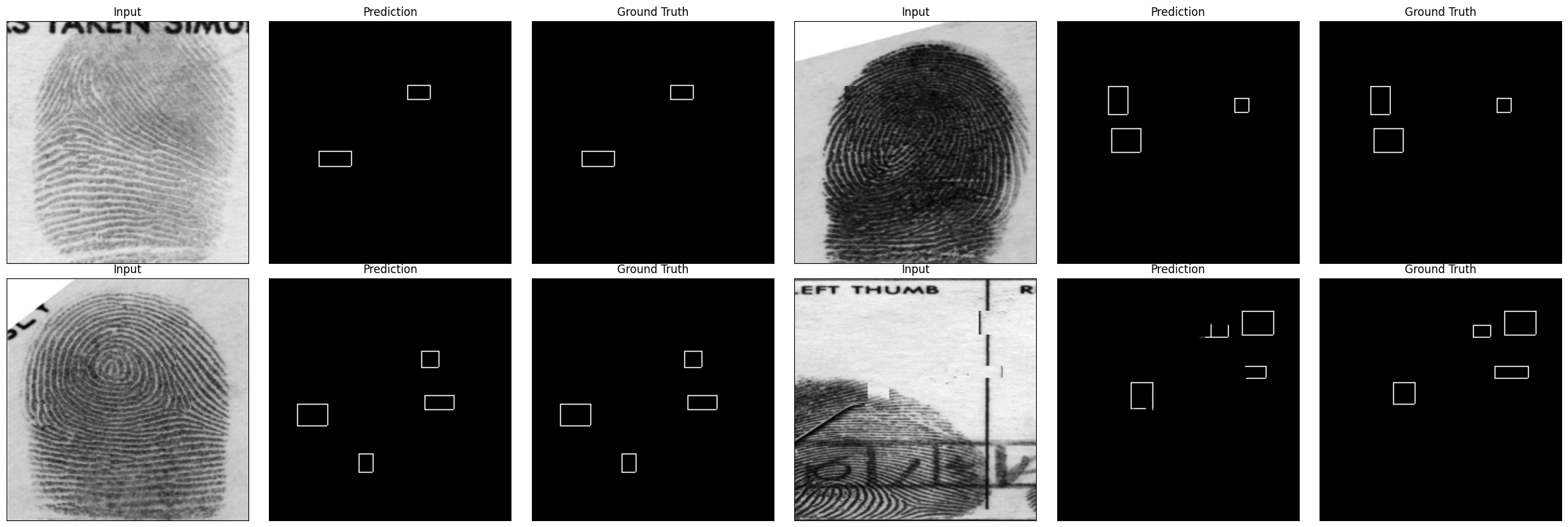}
    \caption{Slap}
    \label{fig:slap_results}
    \end{subfigure}
    \begin{subfigure}{\linewidth}
    \centering
    \includegraphics[width=\linewidth]{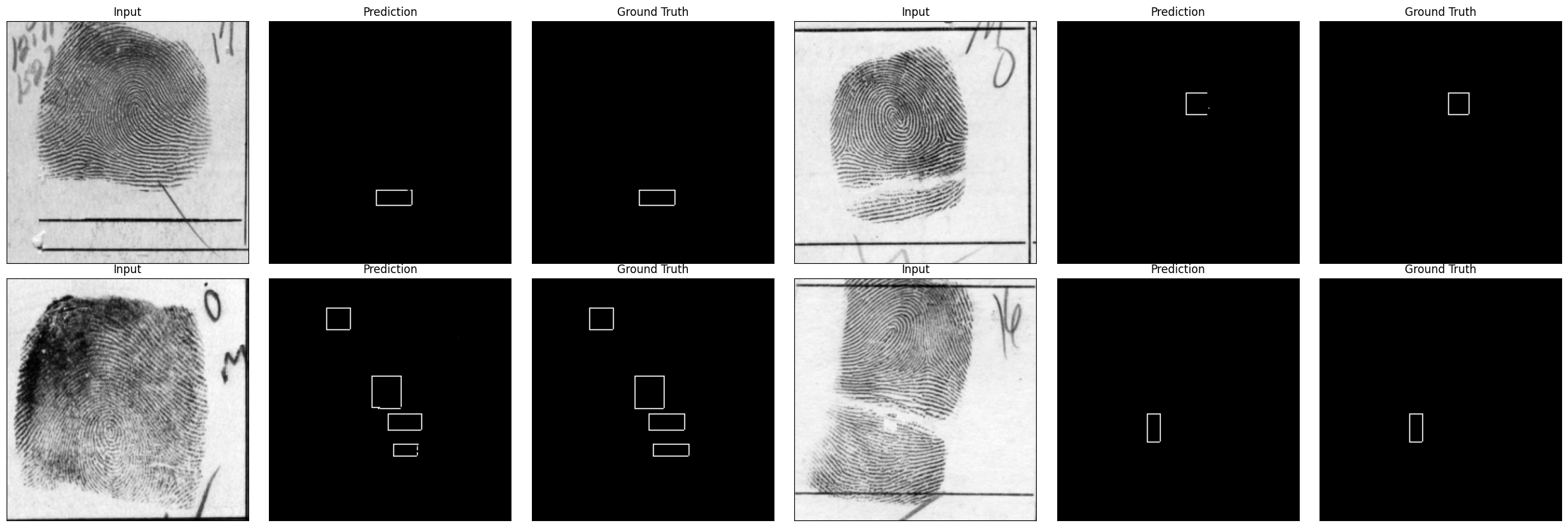}
    \caption{Rolled}
    \label{fig:rolled_results}
    \end{subfigure}
    \caption{Exemplary results with blurred input images for privacy protection.}
    \label{fig:model_results}
\end{figure}

Additionally, we show exemplary results of the CL model on NIST 300a slap data (Fig. \ref{fig:slap_results}) and the NIST 300a rolled data (Fig \ref{fig:rolled_results}) in Figure \ref{fig:model_results}. The leftmost image of the triplets consists of the input image, the second image from the left consists of the model prediction, and the third from the left shows the ground truth.

The distribution of the mosaicking artifact score for the PRD-1-Test, ROD-1 and an additional dataset recorded with a thin-film transistors (TFT) based sensor can be found in the supplementary materials. There, one can see that the score separates into three bands for all datasets. One band with the majority of the samples sitting around zero, one band with only a few examples around the patch weight of 5 and then, the final band with only a handful of outliers at twice the patch weight. The number of examples in the second band around the patch weight score is 2 for the PRD-1-Test dataset. This implies that there is a false matching rate on the PRD-1-Test of 0.061\%.  

The supplementary materials also provide the distribution of the mosaicking artifact score for the ROD-1 dataset analyzed in terms of differences between fingers and also in terms of error types indicated during the acquisition process, such as slipping.

\subsection{Model Robustness}

In order to test the framework's robustness against different data sources, as well as against skin defects that could appear similar to mosaicking artifacts, we conducted a two-fold experiment. First, we trained a second model with the same hyperparameters as the first one, only changing the learning rate and the number of epochs the model was trained on (201 epochs), to adapt to the different data source. In comparison to the first model, which was trained on 245,193 single-shot contactless fingerprint images, the second model was trained on significantly less (26,240) FTIR based pressed fingerprints.  

\begin{figure}[htb]
    \centering
    \begin{subfigure}[t]{0.3\linewidth}
        \includegraphics[width=0.9\linewidth]{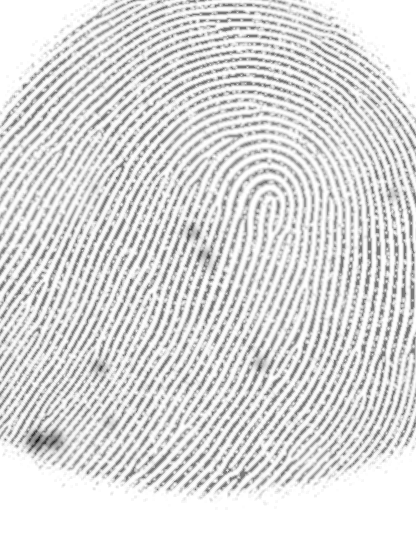}
        \caption{Medium ink variations}
        \label{fig:alteration_ink_medium}
    \end{subfigure}%
    \begin{subfigure}[t]{0.3\linewidth}
        \includegraphics[width=0.9\linewidth]{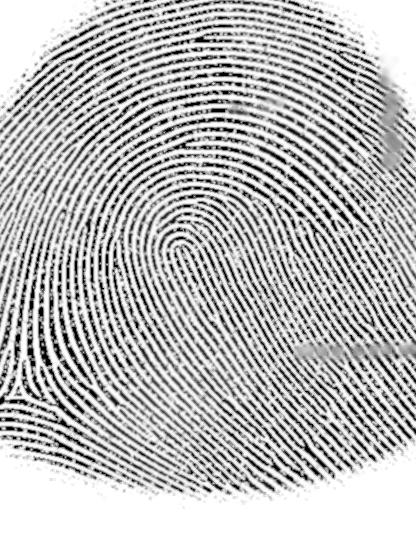}
        \caption{Little scars}
        \label{fig:alteration_scars_little}
    \end{subfigure}%
    \begin{subfigure}[t]{0.3\linewidth}
        \includegraphics[width=0.9\linewidth]{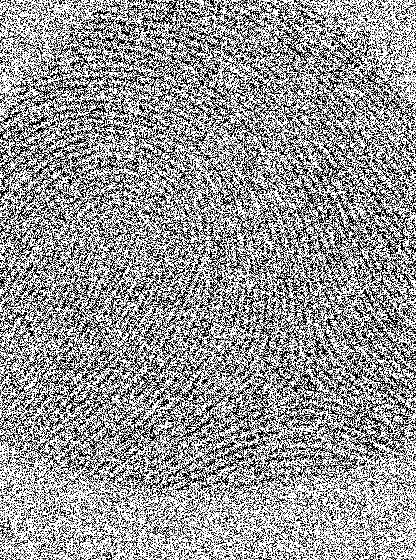}
        \caption{Extensive noise}
        \label{fig:alterations_noise_extensive}
    \end{subfigure}%
    \caption{Synthetic fingerprint alterations based on SynColFinGe applied to SFinGe generated fingerprints.}
    \label{fig:fingerprint_alterations}
\end{figure}

In the next step, we used the fingerprint alteration data. It consists of 100 fingerprint images generated with SFinGe which were modified to show different skin alteration effects. We had the following variations: little noise,  medium noise, extensive noise, little wounds, medium woulds, extensive wounds, little skin problems, medium skin problems, little scars, medium scars, extensive scars, little ink variation, medium ink variation and extensive ink variation. An example can be seen in Figure \ref{fig:fingerprint_alterations}, where \ref{fig:alteration_ink_medium} shows a medium ink variation, \ref{fig:alteration_scars_little} little scars and \ref{fig:alterations_noise_extensive} extensive added image noise. In the supplementary materials, a depiction of all the alteration types is shown. Both models ran inference on the dataset and we used the mosaicking artifact score to measure the extent of false artifact detection. 

Table \ref{tab:fingerprint_alteration_robustness_scores} presents the robustness ablation results for the two models: one trained on contactless images (CL Model) and the other on pressed fingerprint images (PR Model). Each row summarizes the mosaicking artifact scores under varying conditions, represented by the maximum (max), median, mean, and standard deviation (std) scores.

The table columns show the results across multiple types and intensities of image alterations: skin defects (Skin), ink variation (Ink), added noise (Noise), scars (Scar), and wounds (Wounds). Each alteration type has three levels of intensity, denoted by arrows: low ($\downarrow$), medium ($\texttildelow$), and high ($\uparrow$), except skin, where we calculated only low and medium intensities. Additionally, the first column (No) represents the baseline, where no alterations were applied to the images.

Results indicate that, across all alteration types and intensities, the scores for both models remained consistently low, with nearly all values well below the predefined sensitive detection threshold of 5, which corresponds to the patch weight $b$ of equation \ref{equ:mosaick_artifact_score}. A mosaicking artifact score of below the patch weight $b$ indicates that not a single patch was detected. The only exception to the values being above the patch weight is a single, with medium image noise alternated, image for the CL model and three, with medium image noise alternated, images for the PR model. 

\begin{table*}[bth]
    \centering
    \begin{tabular}{cc|c|cc|ccc|ccc|ccc|ccc}
        \toprule
         &  & No & \multicolumn{2}{c|}{Skin} & \multicolumn{3}{c|}{Ink} & \multicolumn{3}{c|}{Noise} & \multicolumn{3}{c|}{Scar} & \multicolumn{3}{c}{Wounds} \\
         & & & $\downarrow$ & \texttildelow & $\uparrow$  & \texttildelow & $\uparrow$ & $\downarrow$ & \texttildelow & $\uparrow$ & $\downarrow$ & \texttildelow & $\uparrow$ & $\downarrow$ & \texttildelow & $\uparrow$ \\
         \midrule
         \multirow{4}{0cm}{\rotatebox{90}{CL Model}} & max & 1.12 & 1.13 & 0.81 & 1.19 & 1.09 & 1.07 & 0.44 & 5.18 & 0.56 & 0.93 & 1.14 & 1.31 & 1.11 & 1.08 & 1.35 \\
         & median & 0.01 & 0.03 & 0.01 & 0.02 & 0.03 & 0.00 & 0.00 & 0.00 & 0.00 & 0.02 & 0.04 & 0.02 & 0.02 & 0.03 & 0.08 \\
         & mean & 0.15 & 0.15 & 0.10 & 0.14 & 0.15 & 0.11 & 0.03 & 0.09 & 0.05 & 0.15 & 0.15 & 0.14 & 0.14 & 0.18 & 0.23 \\
         & std & 0.26 & 0.25 & 0.18 & 0.24 & 0.25 & 0.24 & 0.08 & 0.56 & 0.12 & 0.26 & 0.24 & 0.26 & 0.24 & 0.26 & 0.32\\
         \midrule
         \multirow{4}{0cm}{\rotatebox{90}{PR Model}} & max & 0.00 & 0.00 & 0.00 & 0.00 & 0.00 & 0.01 & 0.00 & 10.41 & 0.13 & 0.00 & 0.00 & 0.00 & 0.00 & 0.00 & 0.21 \\
         & median & 0.00 & 0.00 & 0.00 & 0.00 & 0.00 & 0.00 & 0.00 & 0.00 & 0.00 & 0.00 & 0.00 & 0.00 & 0.00 & 0.00 & 0.00\\
         & mean & 0.00 & 0.00 & 0.00 & 0.00 & 0.00 & 0.00 & 0.00 & 0.24 & 0.00 & 0.00 & 0.00 & 0.00 & 0.00 & 0.00 & 0.00\\
         & std & 0.00 & 0.00 & 0.00 & 0.00 & 0.00 & 0.00 & 0.00 & 1.27 & 0.01 & 0.00 & 0.00 & 0.00 & 0.00 & 0.00 & 0.03\\
         \bottomrule
    \end{tabular}
    \caption{Robustness ablation results of the model trained on contact-less images (CL Model) and contact based images (CB Model). Columns indicate no modification (No), skin damage (Skin), ink problems (Ink), added image noise (Noise), added scars (Scar) and added wounds (Wounds). The arrows indicate low ($\downarrow$), medium ($\texttildelow$) and high ($\uparrow$) intensity of the image modification. The table presents the maximum value (max), median value, mean value and the standard deviation (std) of the mosaick artifact score.}
    \label{tab:fingerprint_alteration_robustness_scores}
\end{table*}

\subsection{Effect of Mosaicking Errors on Equal-Error Rate}

To evaluate the impact of mosaicking errors on fingerprint recognition accuracy, we examined the Equal-Error Rate (EER) using three different Automated Biometric Identification Systems (ABIS): a combination of FingerNet \cite{tang_fingernet_2017} and SourceAFIS \cite{vazan_sourceafis_nodate} for a modern open-source solution, NBIS \cite{watson_users_2007} tools MindTCT and Bozorth3 for an established and fast open-source toolset, and Innovatrics' IDKit \cite{noauthor_idkit_nodate} for a commercial state-of-the-art solution.


FingerNet is a deep learning-based fingerprint segmentation tool that excels in preprocessing fingerprint images, particularly in segmenting and enhancing ridge structures. SourceAFIS is a state-of-the-art (SOTA) open-source fingerprint recognition engine that emphasizes speed and accuracy in matching fingerprint templates. By combining FingerNet for segmentation and SourceAFIS for matching, we used a robust, modern open-source SOTA ABIS that leverages the strengths of both tools \cite{priesnitz_mclfiq_2023}.


The NIST Biometric Image Software (NBIS) suite includes several tools for fingerprint processing and matching. MindTCT is used for minutiae extraction, while Bozorth3 is utilized for minutiae-based fingerprint matching. This combination provides an efficient method for fingerprint recognition that has been widely used in various applications \cite{priesnitz_mclfiq_2023, ruggeri_fingerprint_2020} due to its robustness and speed.


Innovatrics' IDKit represents a commercial off-the-shelf solution known for its high accuracy and reliability in fingerprint recognition. \cite{nist_proprietary_2019}

We categorized mosaicking errors into two groups based on the offset ratio of the artifacts relative to the image dimensions:

\begin{enumerate}
    \item Small Offsets: Offsets between 1\% and 2\% of the input image height and width, rounded towards zero. These represent minor displacements that may still disrupt fingerprint recognition, but are less severe. For a typical fingerprint with a width of 1.5cm, and a height of 2.5cm, this implies an offset of between 2 (0.15mm) and 5 pixels (0.3mm) in the width dimension and 4 (0.25mm) to 9 pixels (0.5mm) in the height dimension. 
    \item Large Offsets: Offsets between 2\% and 7\% of the input image height and width, rounded towards zero. These represent more significant displacements that are likely to have a substantial impact on recognition performance. For the same fingerprint as above, this implies an offset of between 5 (0.5mm) and 20 pixels (1.05mm) in the width dimension and 9 (0.5mm) to 34 pixels (1.75mm) in the height dimension.
\end{enumerate}

By analyzing the EER across these different ABIS and error categories, we aim to understand the robustness of each ABIS in the presence of varying degrees of mosaicking artifacts.

\begin{table}[htb]
    \centering
    \begin{tabular}{l|ccc}
        \toprule
        Artifacts & SourceAFIS [\%] & Bozorth3 [\%] & Idkit [\%] \\
        \midrule
        None &  0.43 & 3.97 & 0.38 \\
        Small Offset & 0.91 & 5.41 & 0.88 \\
        Large Offset & 0.99 & 4.82 & 0.88 \\
        \bottomrule
    \end{tabular}
    \caption{Equal-Error-Rates for artifact free images (first row) calculated using FingerNet for feature extraction and SourceAFIS for matching (SourceAFIS), MindTCT for feature extraction and Bozorth3 for matching (Bozorth3) and Idkit for both feature extraction and matching.}
    \label{tab:eer_stitched_vs_non_stitched}
\end{table}

Table \ref{tab:eer_stitched_vs_non_stitched} summarizes the impact of stitching artifacts on the performance of various ABIS. Both commercial and open-source SOTA systems exhibited substantial increases in EER when comparing artifact-free images to those with stitching artifacts. Specifically, for the SourceAFIS system, the EER increased by 112\% with small pixel offsets and by 130\% with large offsets. Similarly, the Innovatrics Idkit system showed a 132\% increase in EER for both small and large offsets.

\section{Discussion}

\subsection{Model Performance}
\label{subsection:discussion:model_performance}

The models perform best on the modality that they were trained on, which was contactless fingerprint recordings for the CL model and pressed recordings for the PR model. However, the difference in performance of the CL model to slap recordings is only very minute, showing that the model generalized the concept of mosaicking artifacts well and is able to deal with out-of-distribution data. This is even more astonishing as the input data in the NIST 300a dataset is challenging, since it consists of scanned ink prints, which have text or other obstructions in many of the images. Furthermore, the models are able to work with scanned rolled ink fingerprint recordings (NIST 300a rolled) and live scanned TFT based recordings (ROD-1) with only minor decreases in performance when compared to the original modalities.

Furthermore, the findings of the various performance metrics can be put into perspective with comparing the results with the mean mosaicking artifact score difference, as presented in Table \ref{tab:model_performance}. There, the absolute difference of the mosaicking artifact scores between the prediction and the ground truth are averaged over the whole dataset are shown in the rightmost columns, while the traditional computer vision metrics are shown to the left. The correlation between the traditional image based metrics and the mean mosaicking artifact score can be seen. Furthermore, the very low deviation of the prediction from the ground truth in regard to mosaicking artifact score for both the CL and the PR models stays significantly under the detection threshold of the patch weight $b$ of equation \ref{equ:mosaick_artifact_score}, which was set to 5.

As shown via the distribution of the mosaicking artifact score in the supplementary material, the number of false positives is very low. For the PRD-1-Test dataset, the percentage of false matches in only 0.061\%. This shows that the model can be used on large scale databases without the worry that an excessive number of images will be falsely flagged as erroneous.

Furthermore, the clear separation of the result into three bands, which are multiples of the patch weight, as depicted in the supplementary material, indicates a natural clustering and coarse ranking system. 

Despite the promising results, there is room for further optimization, especially concerning real-time processing and deployment on resource-constrained devices. Future work could explore model pruning and quantization techniques to reduce the computational complexity, enabling faster inference speeds. Additionally, training on larger datasets or leveraging more advanced architectures could improve the model's robustness and performance. However, these enhancements must be balanced with the need for efficient deployment, particularly in real-world applications where rapid processing is crucial.

\subsection{Model Robustness}

The results in Table \ref{tab:fingerprint_alteration_robustness_scores} demonstrate that both the CL Model and PR Model exhibit robust performance across various fingerprint alterations, with artifact scores remaining well below the detection threshold of 5 in nearly all scenarios. This consistent performance underscores the effectiveness of both models in distinguishing true mosaicking artifacts from common fingerprint alterations, including skin defects, ink variations, noise, scars, and wounds.

While both models handle alterations effectively, there are minor differences in sensitivity to certain high-intensity conditions. Notably, under medium noise, the CL Model reaches a maximum score of 5.18, which is above the detection threshold and suggesting a slight increase in false-positive risk under image noise conditions. However, a more detailed analysis showed that a value above the detection threshold of 5 was only reached by one example altered with medium image noise, out of the 100 samples with medium image noise presented to the model. This is also reflected in the low mean, std and even zero median scores. In addition, the PR Model also overshot the detection threshold with a maximum of 10.41 for the medium noise case, however, only three images out of the 100 samples with medium image noise presented to the model were above the detection threshold. A plot of the distribution of mosaicking artifact scores for the fingerprint alteration data can be found in the supplementary materials. 

In general, the model shows minimal fluctuation, with near-zero scores across most conditions, while also showing that image noise had the highest impact on model performance. The standard deviation values underscore this robustness, with the PR Model displaying consistently low variability across all types and intensities of alterations, while the CL Model exhibits slightly higher variability, particularly under high noise conditions. Despite these minor differences, both models demonstrate strong robustness and stability, making them well-suited for reliable artifact detection across diverse image conditions. Consequently, we conclude that the proposed framework offers a robust and adaptable solution for training deep learning models to detect mosaicking artifacts, regardless of input data modality or dataset size. 

\subsection{Impact on EER and Dataset Acquisition} \label{subsection:discussion:impact_on_eer_and_dataset_acquisition}

The significant increase in Equal Error Rate (EER) when mosaicking artifacts are present, as shown in Table \ref{tab:eer_stitched_vs_non_stitched}, underscores the importance of detecting and removing such errors during fingerprint acquisition for the performance of ABIS. Our results show that even small artifacts can double the EER, which severely impacts the reliability of fingerprint-based biometric systems. By incorporating an artifact detection model into the fingerprint acquisition process, it is possible to improve the quality of the captured images, thereby reducing EER and enhancing the overall accuracy of biometric identification systems.

This finding highlights the importance of vigilant data acquisition processes, particularly in high-security environments where the accuracy of biometric systems is paramount. While advances in sensor technology have reduced the prevalence of such artifacts, their complete elimination is still not guaranteed. Therefore, implementing a reliable mosaicking artifact detector across all fingerprint sensors and acquisition devices is a prudent measure to ensure the integrity of the captured data. In this context, our model serves a similar function to NFIQ (NIST Fingerprint Image Quality) metrics, offering an additional layer of quality assurance by focusing on artifact detection.

\subsection{Mosaicking Artifact Score}

The proposed mosaicking artifact score introduces an effective method for quantifying fingerprint mosaicking artifacts. The weighting of patch artifacts ($b = 5$) and line artifacts ($c = 0.025$) reflects their relative impact on the mosaicking artifact score. The score itself is validated by EER increases observed in the presence of artifacts, as discussed in section \ref{subsection:discussion:impact_on_eer_and_dataset_acquisition}. 

We propose the detection threshold for counting an observed mosaicking artifact score as a true mosaicking artifact to be linked to the patch weight $b$ of equation \ref{equ:mosaick_artifact_score} for cases where one can expect a patch based misalignment. This requires the model to find at least one closed patch. 

As described in section \ref{subsection:discussion:model_performance}, the mosaicking artifact score provides a sensitive connection between established computer vision metrics like IoU, recall, F1 score, F2 score and accuracy and model performance. This can be seen when comparing the classical metrics presented with the mean mosaicking artifact score difference presented in Table \ref{tab:model_performance}. Therefore, we propose using the mosaicking artifact score for both benchmarking model performance on the validation dataset, as well as for the practical measurement to be used in inference, judging the mosaicking artifact segmentation image in an automated manner.

Further improvements of the mosaicking artifact score could include dynamic weighting based on capture characteristics, integration of additional quality metrics, and adaptation for emerging 3D fingerprint technologies. These enhancements could increase the score’s applicability across varied contexts, while retaining its robustness and reliability in fingerprint quality assessment.

\subsection{Research Impact}

The development of a robust metric for detecting mosaicking artifacts has significant implications for real-world biometric systems. Agencies responsible for national security, law enforcement, and border control could greatly benefit from the integration of such a model into their existing systems. By ensuring that fingerprint images are free from artifacts, these agencies can increase the accuracy and reliability of their biometric databases, reducing the likelihood of misidentification.

Moreover, the metric could be adopted as a standard quality control measure across various biometric acquisition devices. This would not only improve the quality of fingerprint databases but also set a benchmark for future sensor development. The model's ability to generalize across different types of fingerprint data suggests its potential for widespread adoption, providing a consistent method for evaluating and improving the quality of biometric data.

The current standard for assessing fingerprint image quality, NFIQ 2, is primarily designed for pressed fingerprints. However, due to the lack of quality metrics specific to rolled fingerprints, NFIQ 2 is often incorrectly applied to assess these as well. Rolled fingerprints present unique challenges due to variable finger deformation as the finger rolls across the scanner’s sensor surface and because different hardware and software developers implement diverse mosaicking techniques. This mosaicking process can introduce artifacts, which may create new minutiae in the fingerprint or shift the position of existing ones. These artifacts pose significant risks in identification, potentially leading to misidentification or missed matches. Central fingerprint registers, which rely on artifact-free reference and comparison samples, are particularly vulnerable to these issues. Consequently, tools that detect mosaicking artifacts at all stages of processing could enhance the reliability of identification systems based on rolled fingerprints, starting with quality control at the initial fingerprint acquisition stage—such as in asylum applications or identity verification services. Currently, human operators are often required to manually assess fingerprint quality and suitability for inclusion in central registers, underscoring the need for improved automated quality assessment methods.

\subsection{Outlook}

Looking ahead, one promising avenue for further research is the supervised finetuning of self-supervised pretrained models on real fingerprint mosaicking artifacts. While our self-supervised approach has proven effective, training on real-world artifacts could bridge the gap between simulated and actual data. However, this would require the collection of a large, diverse dataset of annotated real artifacts from multiple sensors and users, which poses significant logistical challenges.

Another research avenue is the integration of the model into edge devices or directly into fingerprint sensors. By embedding the artifact detection capability at the point of data acquisition, it would be possible to detect errors in real-time, improving the quality of the data before it reaches the processing stage.

Recently, discussions started on improving the NFIQ 2 standard in terms of including contactless fingerprints and rolled fingerprints as types of fingerprints officially supported by the NFIQ. Therefore, our approach may contribute with its framework to be considered for image quality assessment, especially for the important detection and correction of mosaicking and related artifacts in fingerprints.

Advancements in technology may enable contactless fingerprint in combination with capturing multiple images from different angles to achieve interoperability with present rolled fingerprints. Mosaicking will also play an important role for this kind of future application and fingerprint acquisition methods.   

In conclusion, while our model has demonstrated impressive capabilities, there are still opportunities for further improvement and application. Continued research in this area could lead to significant advancements in the accuracy and reliability of biometric systems, ultimately enhancing security and efficiency in various applications.
\section{Conclusion}
In this paper, we have presented a novel deep learning-based framework to detect and classify mosaicking artifacts in fingerprint images. Our proposed method leverages a self-supervised learning paradigm, enabling us to train the model on a large dataset of unlabeled fingerprint images without the need for manual annotation.

By introducing a novel mosaicking artifact score, we have provided a quantitative measure to assess the severity of mosaicking errors. This score enables automated evaluation and prioritization of fingerprint images for further processing or rejection.

Our experimental results demonstrate the effectiveness of our approach in accurately detecting and classifying hard mosaicking artifacts. The model exhibits robust performance across different fingerprint modalities, including contactless, rolled, and pressed fingerprints.

The successful application of our method has the potential to significantly improve the quality and reliability of fingerprint-based biometric systems. By identifying and mitigating mosaicking artifacts, we can enhance the accuracy of fingerprint matching and verification, leading to more secure and efficient biometric authentication.

Future work may involve exploring real-time implementation strategies for practical applications and the extension of the approach to real mosaicking artifact data. 

\ifCLASSOPTIONcaptionsoff
  \newpage
\fi

\printbibliography

\end{document}